%% file: arxiv.tex
\documentclass[11pt, a4paper, logo]{googledeepmind}

\pdftrailerid{redacted}

\makeatletter
\renewcommand\bibentry[1]{\nocite{#1}{\frenchspacing\@nameuse{BR@r@#1\@extra@b@citeb}}}
\makeatother

\usepackage{kantlipsum, lipsum}
\usepackage{dsfont}
\usepackage{gdm-colors}

\usepackage{thmtools, thm-restate}
\usepackage{proof-at-the-end}

\usepackage{algorithm}
\usepackage{algorithmic}

\def\data{\mathcal{D}}

\def\1{\mathbbm{1}}

\def\hat{\widehat}

\DeclareMathOperator*{\argmax}{arg\,max}

\definecolor{red}{RGB}{255,0,0}
\definecolor{blue}{RGB}{0,0,255}
\definecolor{green}{RGB}{0,255,0}
\definecolor{orange}{RGB}{255,165,0}
\definecolor{purple}{RGB}{128,0,128}
\definecolor{teal}{RGB}{0,128,128}

\newcount\Comments
\newcommand{\kibitz}[2]{\ifnum\Comments=1{\textcolor{#1}{\textsf{\footnotesize #2}}}\fi}

\newcommand{\smallllm}{Gemini Nano}
\newcommand{\bigllm}{Gemini Pro}

\usepackage{fancyvrb}

\usepackage[authoryear, sort&compress, round]{natbib}

\graphicspath{{figures/}}

\title{Efficient Exploration for LLMs}

\reportnumber{} % Leave blank if n/a

\renewcommand{\today}{}

\author[1]{Vikranth Dwaracherla}
\author[1]{Seyed Mohammad Asghari}
\author[1]{Botao Hao}
\author[1, 2]{Benjamin Van Roy}

\affil[1]{Google DeepMind}
\affil[1,2]{Stanford University}

\begin{abstract}
We present evidence of substantial benefit from efficient exploration in gathering human feedback to improve large language models.  In our experiments, an agent sequentially generates queries while fitting a reward model to the feedback received.  Our best-performing agent generates queries using double Thompson sampling, with uncertainty represented by an epistemic neural network.  Our results demonstrate that efficient exploration enables high levels of performance with far fewer queries.  Further, both uncertainty estimation and the choice of exploration scheme play critical roles.
\end{abstract}

\begin{document}

\maketitle

% Incude paper content from external files
\input{arxiv_content}

% Bibliography components
\bibliographystyle{abbrvnat}
\nobibliography*
\bibliography{references}

\newpage
% If you add a bibtex entry of your own paper (this paper), you can
% show its full citation inline using \citeas, as above. Note that
% this citation removes the trailing full stop. To make \citeas work,
% you need to load the bibliography data. This can be done in two
% ways:
%
%    1. If you already have a printed bibliography with \bibliography{...},
%       then add the command "\nobibliography*", no arguments, before that.
%    2. If you don't otherwise print a bibliography, add the command
%       \nobibliography{...} at the end of your document.

\appendix

\section{Human Preference Simulator}
\label{app:simulator}

We use an oracle reward model to construct our preference simulator. Given a query, comprising a prompt and two potential responses, the preference simulator produces binary feedback indicating a preference between the two responses by first computing scores for each of the two responses. To simulate preference probabilities from these scores, we employ the Bradley-Terry model \cite{bradley1952rank}, a well-established methodology in decision analysis. 
While we make use of binary feedback sampled from the preference probabilities in the training pipeline, we directly use the preference probabilities in the assessment pipeline. The direct use of preference probabilities in our assessment pipeline is to reduce stochasticity in evaluation. 

The oracle reward model itself is constructed with a two-part architecture, featuring a torso responsible for producing embeddings and a linear layer head that generates scalar estimates. The torso is initialized to the pre-trained \bigllm{} transformer torso, while the linear head uses Xavier initialization \cite{glorot2010understanding}. The training process involves optimizing both the torso and reward head  through cross-entropy loss function applied to the Anthropic Helpfulness and Harmlessness datasets \cite{bai2022training}. 

Our oracle reward model attains an accuracy of $0.755$ on the Anthropic Helpfulness and $0.748$ on the Anthropic Harmlessness eval datasets. Notably, this level of performance is higher than the performance of the largest models reported in \citep{bai2022training}. 

Our feedback simulator is designed to capture uncertainty stemming from both variations among different raters and within the responses of individual raters. However, we believe that the predominant source of uncertainty arises from the differences among raters. This is because it is unlikely that the same rater, when presented with the same query several times, would indicate inconsistent preferences. In our preference simulator, we've observed that the average probability of obtaining the same label twice for a query is approximately 0.62, with a standard deviation of 0.14 across various queries. This suggests that preference labels for a batch of queries can be thought of as being generated by different raters. This means that when trained on preference data labeled using our preference simulator, it is akin to optimizing for preferences of a pool of raters rather than preferences of a single rater.

Note that, since \bigllm{} is far larger than \smallllm{}, choices are made by a much more complex model than that available to the agent.  This difference in scale is intended to reflect the fact that humans may exhibit more complex behavior than that modeled by the agent.

\section{Hyperparameter Selection for Experiments} \label{app:hypers}

We tune the hyperparameters of our agent to optimize performance. These hyperparameters include the l2 regularization strength, learning rate, and the number of stochastic gradient descent (SGD) steps executed after each epoch of interaction. Every stochastic gradient descent (SGD) step in our training process is executed on a batch of data randomly sampled from the agent's replay buffer. 

We sweep the learning rate over $\{1e-6, 1e-5, 1e-4\}$ for both point estimate and ENN reward models and pick the best learning rate. We found that the best learning rate is consistent across both our joint nll experiments described in Section \ref{sec:joint_nll} and our active learning experiments. 

To adapt to the evolving nature of the data collection process, we implement a strategy of decay for the regularization strength. The regularization strength, denoted as $\lambda$ in Equations \ref{eq:point_loss} and \ref{eq:enn_loss}, is adjusted in accordance with the volume of data accumulated by the agent. Specifically, for each stochastic gradient descent (SGD) step carried out at every epoch on a batch comprising $B=32$ preference data points from the replay buffer, we incorporate a decayed regularization strength given by $\lambda = \frac{32 \times \lambda'}{|\data|}$, where $\data$ represents the total number of feedback data points amassed by the agent, and we tune $\lambda'$ by sweeping across $\{0.1, 1, 10, 100, 1000\}$ for all the agents.

We also swept over the number of sgd steps performed after each epoch of interaction from $\{1, 10, 100\}$. We observed that \texttt{infomax} agent benefits from running for more sgd steps while the performance of other agents deteriorates beyond a point due to over fitting. 

In the case of ENN models, we also tune the output scale parameter, responsible for regulating the diversity of ensemble particles. In specific, we  sweep over values $\{0.1, 1, 10\}$ and pick the value which leads to best performance per agent. Note that we use weight regularization towards initial weights similar to Section 2.1 in \citet{dwaracherla2020hypermodels}.

Futhermore, we also tuned the number of hidden units for a two-layer MLP in point estimate model by sweeping over $\{128, 256, 512, 1024, 2048\}$ in the context of our uncertainty estimation experiments detailed in Section \ref{sec:joint_nll}. Despite our thorough exploration, we observed no substantial enhancement in performance associated with an increase in hidden units. Consequently, we opted to employ $128$ hidden units for MLPs across all of our experimental results presented in this paper. 

\section{Performance across different ensemble sizes}

\begin{figure}[!ht]
    \centering
    \includegraphics[width=0.6\columnwidth]{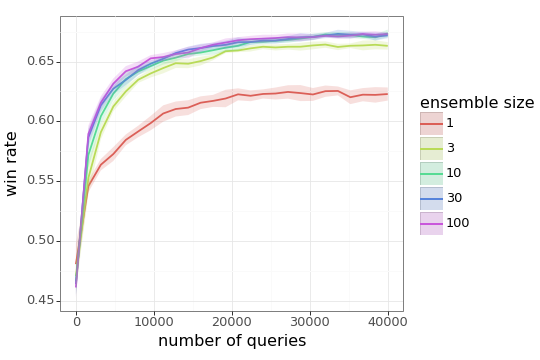}
    \caption{Performance across different ensemble sizes}
    \label{fig:ensemble_sweep}
\end{figure}

Figure \ref{fig:ensemble_sweep} shows the performance of {\tt double Thompson sampling} agent across various ensemble sizes. We observe that performance improves with increasing ensemble size; however, beyond a certain threshold, the improvements become marginal. Specifically, the improvement plateau around an ensemble size of 10. Therefore, we use an ensemble size of 10 in our experiments.

\section{Exploration Algorithms with a Single Point Estimate} \label{app:point}

\begin{figure}[!ht]
    \centering
    \includegraphics[width=0.6\columnwidth]{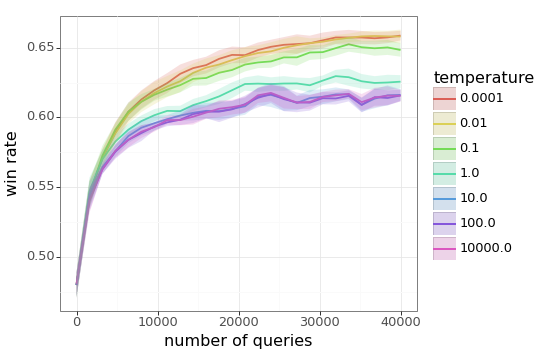}
    \caption{This plot demonstrates the performance of Boltzmann agent across different temperatures.}
    \label{fig:boltzmann_winrate}
\end{figure}

In this section, we evaluate the performance of the {\tt Boltzmann} algorithm across various temperature values. We vary the temperature parameter, denoted as $\tau$, in the Boltzmann exploration scheme (refer to Algorithm \ref{alg:Boltzmann}). The range of temperatures explored includes $\tau \in \{1e-4, 1e-2, 1e-1, 0, 1, 10, 100, 1000\}$. The corresponding performance of the {\tt Boltzmann} agent under different $\tau$ values is illustrated in Figure \ref{fig:boltzmann_winrate}.Notably, we observe optimal performance for {\tt Boltzmann} agents with smaller temperatures. Additionally, our findings affirm expectations that {\tt Boltzmann} agents with very high temperatures exhibit performance akin to a passive agent.

We additionally experimented with a variant of the {\tt Boltzmann} scheme known as {\tt Greedy-Boltzmann}, as outlined in Algorithm \ref{alg:greedy_boltz}. In the {\tt Greedy-Boltzmann} approach, one response is chosen greedily, relying on the reward model, while the selection of the other response follows the Boltzmann exploration scheme. Conceptually, {\tt Greedy-Boltzmann} can be viewed as a modification of Boltzmann with two temperatures, denoted as $\tau_1$ and $\tau_2$, where $\tau_1$ is fixed at 0, and $\tau_2$ is subject to variation.

\begin{algorithm}
\caption{Greedy-Boltzmann}
\label{alg:greedy_boltz}
\textbf{input:} $x$, $\pi$, $r$ \\
\textbf{hyperparams:} $\tau$, $N$
\begin{algorithmic}[1]
\STATE sample responses $\tilde{y}_1,\ldots,\tilde{y}_N \sim \pi(\cdot|x)$
\STATE select response $i \in \argmax_n r(x, y_n)$ 
\STATE probs $q_n = \frac{\exp(r(x,\tilde{y}_n) / \tau)}{\sum_{n'=1, n \neq i}^N \exp(r(x,\tilde{y}_{n'}) / \tau)} \forall n \neq i$, $q_i = 0$
\STATE sample $i' \sim q$
\end{algorithmic}
\textbf{return} $y_i,y_{i'}$
\end{algorithm}

\begin{figure}[!ht]
    \centering
    \includegraphics[width=0.6\columnwidth]{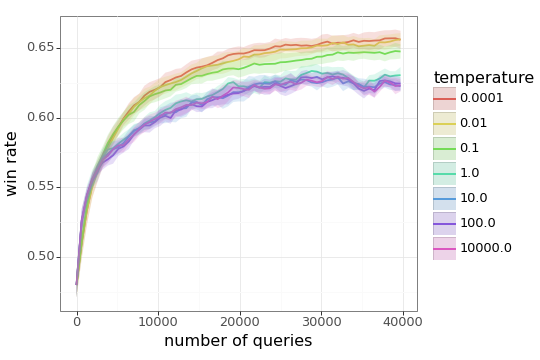}
    \caption{Performance of Greedy-Boltzmann across different temperatures for Boltzmann. We can see that best  Greedy-Boltzmann and best Boltzmann agent perform very similarly, and Greedy-Boltzmann doesn't offer an advantage over Boltzmann.}
    \label{fig:greedy-boltzmann}
\end{figure}

The performance of the {\tt Greedy-Boltzmann} variant across different temperatures is illustrated in Figure \ref{fig:greedy-boltzmann}. Notably, after fine-tuning the temperature parameter, the results indicate that {\tt Greedy-Boltzmann} doesn't improve over the performance achieved by the standard {\tt Boltzmann} agent, as presented in Algorithm \ref{alg:Boltzmann}. 

The {\tt Boltzmann} and {\tt Greedy-Boltzmann} agents can be conceptualized as approximating various exploration strategies determined by the temperatures used in Algorithms \ref{alg:Boltzmann} and \ref{alg:greedy_boltz}. This encompasses examples such as "greedy" exploration, involving the selection of the two best responses; "greedy-uniform" exploration, where the first response is chosen greedily and the second is randomly selected; and "passive" exploration, where both responses are sampled randomly. Therefore, when evaluating the performance of {\tt Boltzmann} and {\tt Greedy-Boltzmann}, we are essentially assessing a broad spectrum of exploration schemes derived from a point estimate reward model.

\section{Dyadic NLL}

To assess the quality of joint predictions,  we sample batches of queries $(x_1, . . . , x_\tau )$ and evaluate the log-loss with respect to their corresponding preference probabilities $(p_1, . . . , p_\tau )$. In high-dimensional input spaces, queries sampled uniformly at random are typically nearly independent. Consequently, a significantly large value of $\tau$ becomes necessary to effectively discern joint predictions that capture the interdependencies among queries. However, this approach quickly becomes impractical due to the exponential growth in computational requirements as $\tau$ increases.

To address this challenge, dyadic sampling, as proposed in \citep{osband2022evaluating}, presents a pragmatic heuristic. Dyadic sampling strategically selects queries to ensure that the distinguishing power of effective agents can be achieved with a manageable $\tau$. This method involves initially sampling two queries and then repeatedly selecting $\tau$ queries with replacement from this pair. Subsequently, the joint negative log-likelihood (nll) is computed over the sampled queries using their respective preference probabilities. This process is iterated several times, and the average joint nll is reported as the dyadic joint nll.

We utilized the code in the enn library \citep{osband2023epistemic} to implement dyadic joint nll. 
\end{document}

%% file: arxiv_content.tex
% A few local macros that are used by the example content.
\newcommand{\expect}[2]{\mathds{E}_{{#1}} \left[ {#2} \right]}
\newcommand{\myvec}[1]{\boldsymbol{#1}}
\newcommand{\myvecsym}[1]{\boldsymbol{#1}}
\newcommand{\vx}{\myvec{x}}
\newcommand{\vy}{\myvec{y}}
\newcommand{\vz}{\myvec{z}}
\newcommand{\vtheta}{\myvecsym{\theta}}

\section{Introduction}
\label{se:introduction}

Large language models demonstrate remarkable capabilities after learning from enormous volumes of text data \cite{anil2023palm,NEURIPS2022_c1e2faff,openai2023gpt4}.  Yet, reinforcement learning from human feedback (RLHF) greatly improves their behavior even after only tens of thousands of interactions \cite{stiennon2020learning,bai2022training,ouyang2022training,glaese2022improving}.
The uptake of chatbots affords opportunities to gather increasing volumes of human feedback, with each engagement eliciting expressions of satisfaction or preference \cite{openai2022chatgpt}.  
It is natural to wonder what new capabilities may emerge with this growing source of data.  Superhuman ingenuity remains an alluring possibility.  

With increasing volumes, more can be inferred from human feedback.  This allows behavior to deviate further from that of a pretrained model. But given that this process learns only from humans, how can we hope for emergence of superhuman ingenuity?  Perhaps such an outcome is plausible because rating is easier than synthesizing novel content.  This is analogous to how, for an NP-complete problem, though solution is hard, verification of a proposed solution is easy.

Suppose, for example, a pretrained model extrapolates from its training data to generate large numbers -- perhaps millions or billions -- of ideas, one of which is ingenious.  While a human may not have come up with that idea, learning from enough human feedback can identify it from among the large number of ideas generated by the model.  And, building on this innovation, further extrapolation can continue to expand the frontier of ingenuity.  In this way, with enough human feedback, a model ought to become capable of generating content that a human could not. But will gathering the required feedback take months, years, or decades?

We present in this paper evidence of enormous benefit to active exploration.  By {\it active exploration} we mean the tailoring of interactions to elicit useful feedback.  In particular, our results demonstrate that high levels of performance can be attained with far less feedback.  This acceleration may enable superhuman ingenuity much, perhaps decades, sooner.

A common practice in reinforcement learning from human feedback (RLHF) is to send queries, each comprised of a prompt and a pair of distinct responses, to human raters.  Each rater expresses a preference for one response over the other.  Prompts are drawn from a corpus, while responses are generated by the large language model.  As this process progresses, a reward model is fit to the data and steers subsequent responses to align with with feedback received thus far.

In this paper, we restrict attention to the aforementioned sort of interaction, in which each query includes a prompt and pair of distinct responses.  We refer to the standard practice of sampling each pair of responses using the language model as {\it passive exploration}.  We compare the performance of passive exploration to several active exploration algorithms.  One is Boltzmann exploration, which tends to select responses with higher predicted reward.  We also tried two approaches that leverage uncertainty estimates offered by an epistemic neural network (ENN).  The first, which we refer to as {\it infomax}, selects a pair of responses with an aim of maximizing information revealed by the feedback.  This belongs within the widely used collection of algorithms that aim to maximize information gain (see, e.g., \cite{mackay1992information,Sun2011,Houthooft2016,sadigh2018planning}).  The second, called {\it double Thompson sampling} \cite{NIPS2016_9de6d14f}, samples responses according to the probability they are optimal.  These exploration algorithms will be described more precisely in Section \ref{sec:exploration-algorithms}.

\begin{figure}[!ht]
\centering
\includegraphics[width=0.6\columnwidth]{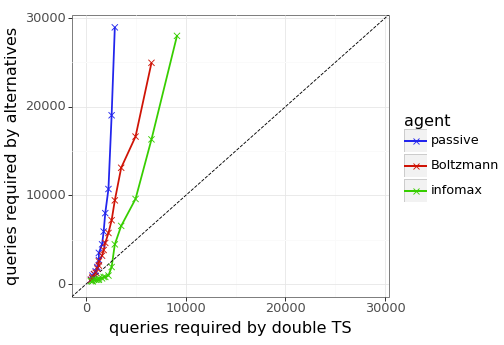}
\caption{Queries required by double TS versus alternatives to attain various levels of performance.}
\label{fig:active-vs-passive}
\end{figure}

Figure \ref{fig:active-vs-passive} compares empirical results produced using different exploration algorithms.  The experiments that generated these results are described in Section \ref{se:results}.  Each plotted point corresponds to a level of performance attained.  The horizontal coordinate identifies the number of queries required by double TS to reach that performance level, while the vertical coordinate identifies that required by an alternative.  The plot for passive exploration clearly demonstrates that {\bf active exploration using double TS greatly reduces the number of queries required to reach high levels of performance}.  Boltzmann exploration performed best among algorithms we tried that used only a point estimate reward model, without uncertainty estimates.  The plot for Boltzmann demonstrates that {\bf uncertainty estimates, as used by double TS, enable dramatic improvement}.  Finally, the plot for infomax shows how, even among tried and tested algorithms that leverage uncertainty estimates, {\bf the choice of exploration algorithm can drive large performance differences}. 

While these are to our knowledge {\bf the first results demonstrating substantial benefits from active exploration in tuning large language models}, they build on a long history of work pertaining to exploration algorithms \cite{lattimore2020bandit}.  In particular, our problem is an instance of the contextual dueling bandit \cite{yue2012k, dudik2015contextual, saha2021optimal} and our algorithms build on information-seeking schemes \cite{mackay1992information,Sun2011,hennig2012entropy, ryzhov2012knowledge,russo2014learning,Houthooft2016,sadigh2018planning} and Thompson sampling \cite{thompson1933likelihood,russo2018tutorial,NIPS2016_9de6d14f}.  Further, our effort continues a line of work that has scaled efficient exploration algorithms to increasingly complex environments using neural networks \cite{bellemare2016unifying, osband2016deep,lu2017ensemble, ostrovski2017count, riquelme2018deep, burda2018exploration, osband2019deep, zhou2020neural, zhang2020neural, dwaracherla2020hypermodels, badia2020never, pmlr-v216-osband23a}.

\section{Experimentation Pipeline} \label{sec:pipeline}

We start by presenting the experimentation pipeline we use to study exploration algorithms.  This pipeline builds on existing tools, including the Anthropic datasets \cite{bai2022training} and the \smallllm{} and \bigllm{} pretrained language models \cite{geminiteam2023gemini}.  It makes use of a human feedback simulator, which generates in response to each query a binary expression of preference between responses.  The pipeline is made up of two parts: a learning pipeline and an assessment pipeline.  The former governs the interface between the agent and the human feedback simulator in the process of sequential querying and learning.  The latter governs the interface between the pretrained language model, the new response generation model, and the human feedback simulator in the process of assessing relative performance.

\begin{figure*}[!ht]
    \centering
    \includegraphics[width=0.8\textwidth]{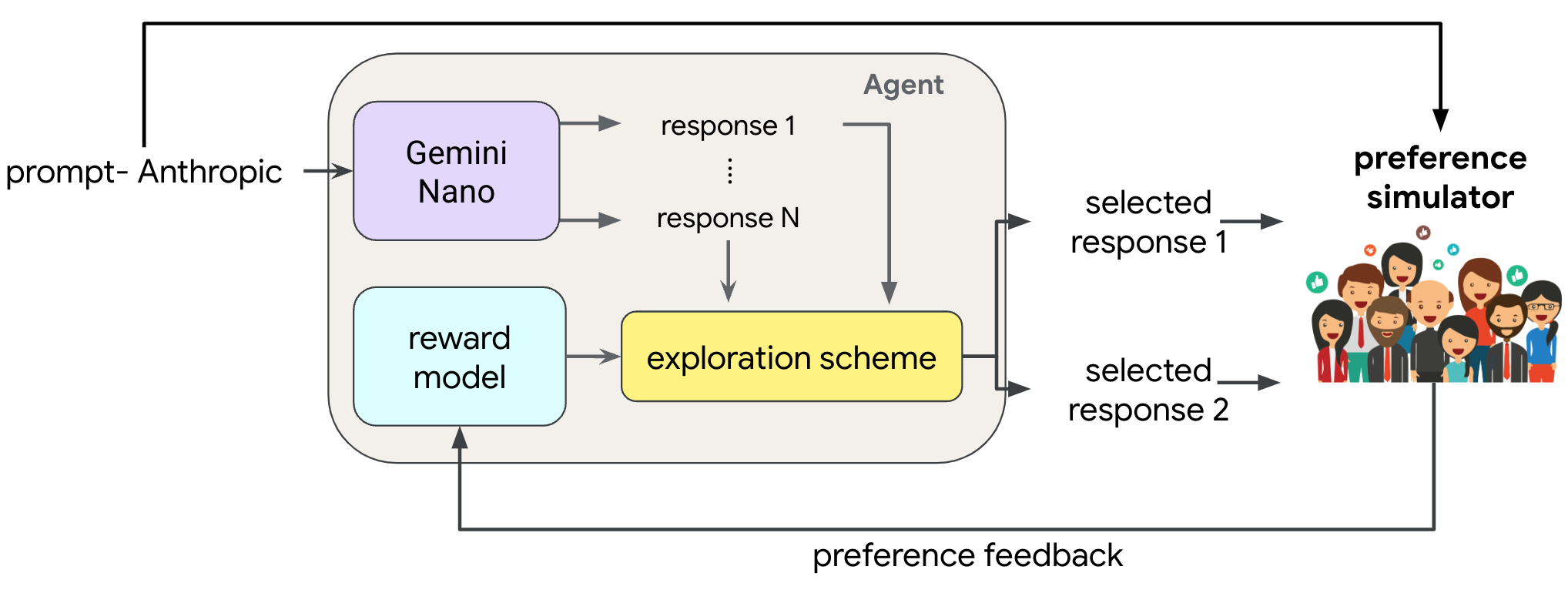}
    \caption{The sequential querying and learning pipeline.}
    \label{fig:RLHF-pipeline}
\end{figure*}

An agent learns sequentially from feedback to queries, each comprised of a prompt and two alternative responses.  As illustrated in Figure \ref{fig:RLHF-pipeline}, each query is crafted by the agent and presented to a human preference simulator, which indicates a binary preference between the two.  Over each epoch of interaction, the agent transmits a batch of $B$ queries and receives the $B$ bits of feedback.  Each prompt is sampled uniformly from the Anthropic Helpfulness Base train dataset.  Each agent we study, when presented with a prompt, crafts its pair of responses by first generating $N$ candidates using the \smallllm{} model and then applying an exploration algorithm that selects two from among these $N$.  The exploration scheme accesses a reward model which is trained on queries and feedback observed thus far.  Each agent we consider is distinguished by its exploration algorithm and the architecture and training algorithm that produce its reward model.  In some of the agents we consider, the reward model takes the form of an epistemic neural network, which offers the exploration algorithm access to uncertainty estimates in addition to point estimates of reward.  Each reward model builds on the torso of the \smallllm{} model.  By this we mean that the reward model first computes the last-layer embedding of the pretrained transformer model and then applies an multilayer perceptron (MLP) head.  We elaborate on architectures and training algorithms in Section \ref{sec:reward-learning-algorithms}.

To simulate how humans choose between responses, we use a reward model that scores each prompt-response pair.  For each query, a preference is sampled according to the Bradley-Terry choice model based on scores assigned to the two prompt-response pairings.  The reward model used by this simulator is fit to the Anthropic datasets, with an architecture that reuses the torso of the \bigllm{} language model.  Further detail is provided in Appendix \ref{app:simulator}.  Note that, since \bigllm{} is far larger than \smallllm{}, choices are made by a much more complex model than that available to the agent.  This difference in scale is intended to reflect the fact that humans may exhibit more complex behavior than that modeled by the agent.  Algorithm \ref{alg:learning-pipeline} offers a concise presentation of interactions -- in particular, what is transmitted (tx) and received (rx) to and from the agent and simulator -- in our learning pipeline.

\begin{algorithm}
\caption{learning interface}
\label{alg:learning-pipeline}
\textbf{input:} prompt\_set, agent, feedback\_simulator \\
\textbf{hyperparams:} $B, T$
\begin{algorithmic}[1]
\FOR{$t$ in $1,\ldots,T$}
\STATE tx to agent: $B$ prompts
\STATE rx from agent: two responses per prompt
\STATE tx to simulator: $B$ queries
\STATE rx from simulator: $B$ bits of feedback
\STATE tx to agent: $B$ bits of feedback
\ENDFOR
\end{algorithmic}
\end{algorithm}

\begin{figure*}[!ht]
    \centering
    \includegraphics[width=0.8\textwidth]{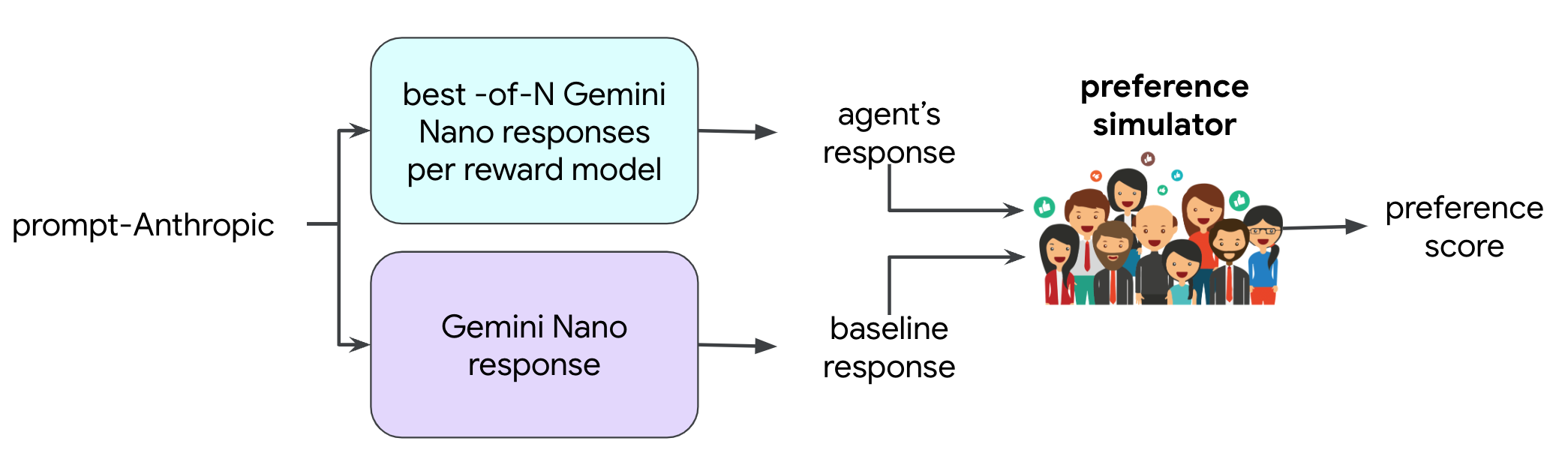}
    \caption{The performance assessment pipeline.}
    \label{fig:performance-pipeline}
\end{figure*}

Figure \ref{fig:performance-pipeline} illustrates our pipeline for assessing agent performance.  Performance is measured relative to the \smallllm{} model.  A sequence of prompts is sampled from Anthropic Helpfulness Base eval dataset.  For each, two responses are sampled.  One by \smallllm{} and the other by a new response generation model that uses the learned reward model.  This new model operates by sampling $N$ responses using \smallllm{} and then selecting the one that scores highest according to the agent's reward model.  The human preference simulator outputs its probability of choosing the agent's response over the alternative generated by \smallllm{}.  These probabilities are averaged over prompts, and this average is referred to as the agent's {\it win rate}, as it represents the fraction of time that the agent's response is preferred.  Note that the win rate can also be estimated by averaging binary indications of simulated choice, though a larger number of queries would be required for an estimate produced in this manner to converge.  Algorithm \ref{alg:assessment-pipeline} offers a concise presentation of interactions in the assessment phase.

\begin{algorithm}
\caption{assessment interface}
\label{alg:assessment-pipeline}
\textbf{input:} prompt\_set, model1, model2, feedback\_simulator

\begin{algorithmic}[1]
\FOR{prompt in prompt\_set}
\STATE tx to models: prompt
\STATE rx from models: one response per model
\STATE tx to simulator: query (prompt + 2 responses)
\STATE rx from simulator: prob of preferring response 1
\ENDFOR
\end{algorithmic}
{\bf return} average across preference probabilities
\end{algorithm}

Note that our experiment pipeline sidesteps the sort of policy-gradient methods typically used to optimize reward.  Instead, our agent samples $N$ responses from the base language model (\smallllm{}) and selects from among those the one that maximizes reward.  This best-of-$N$ procedure serves to approximate policy-gradient-based optimization, but without its cumbersome computational requirements.  The best-of-$N$ procedure also cultivates more transparent analyses, since it avoids poorly understood dependence on the hyperparameter tinkering often required to obtain reasonable results from policy gradient methods.  A prototypical policy gradient approach minimizes a loss function that balances between two objectives: similarity to the base language model and alignment with reward.  A scalar hyperparameter multiplies the similarity measure, striking the balance between these objectives.  The parameter $N$ plays a similar role in the best-of-$N$ approach.  As $N$ increases, maximizing over responses more closely aligns the agent with reward.  Moderating $N$ encourages agent behavior more similar to the base language model.

\section{Reward Model Architectures and Training}
\label{sec:reward-learning-algorithms}

Reward models guide response selection in both the learning and assessment phases of our experiment pipeline.  We consider two types of reward models, each of which is fit to observed preference data.  The first is a point estimate that assigns a reward to each prompt-response pair.  The second depends additionally on an epistemic index.  Sampling an epistemic index from a reference distribution induces randomness in reward, which models epistemic uncertainty about the reward.  In this section, we describe the neural network architectures and training algorithms used in our experiments.

\begin{figure*}[!ht]
    \centering
    \includegraphics[width=0.8\textwidth]{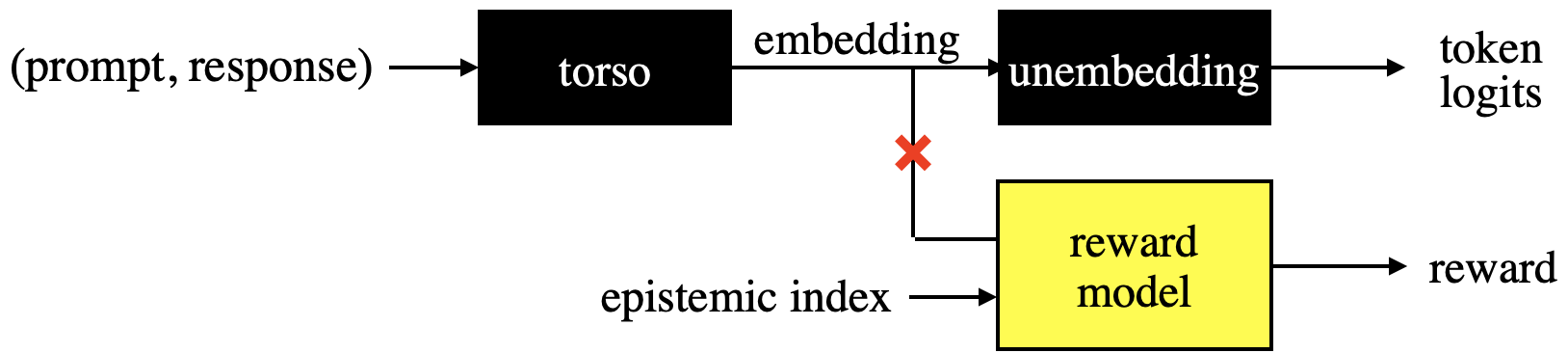}
    \caption{Our reward models take as input the last-layer embedding of the \smallllm{} language model.  A stop gradient prevents torso updating of torso weights.}
    \label{fig:reward_model}
\end{figure*}

We train reward models that each take as input the last-layer embedding of the \smallllm{} language model.  As illustrated in Figure \ref{fig:reward_model}, a reward is assigned to a prompt-response pair by first passing it through the language model torso and then through a reward model.

% We consider two flavors of reward models.  The first is a point estimate. 
% % represented by a \ben{describe neural network architecture that we use}.
% The second uses an epistemic neural network (ENN) to represent epistemic uncertainty about reward.  An ENN takes an epistemic index as an additional input.
% % \ben{describe ENN architecture that we use}

\subsection{Point Estimate}\label{sec:point}

In our architecture, a point estimate reward model takes the form of a feedforward multi-layer perceptron (MLP).  This reward model takes as input the last-layer embedding of the \smallllm{} language model, which itself takes as input a prompt-response pair $(x,y)$.  The reward model then outputs a scalar reward $\hat r_\theta(x,y)$.  Here, $\theta$ is the vector of MLP parameters.

We train reward models on preference data.  Each data point consists of a query, consisting of a prompt and pair of responses, and a binary indication of preference between the responses.  Given a set $\data$ of such data points, to compute MLP parameters, we optimize the loss function
\begin{equation}\label{eq:point_loss}
\mathcal{L}_{\rm point}(\theta | \data) = \sum_{(x,y,y',c) \in \data} \mathrm{ce}(r_\theta(x,y), r_\theta(x,y'),c) + \lambda \|\theta\|_2^2,
\end{equation}
where $\lambda$ is the regularization strength, $c$ indicates choice or preference, and $\mathrm{ce}(\cdot, \cdot, \cdot)$ denotes the cross entropy loss:
\begin{equation}
    \mathrm{ce}(R, R', c) = - (1-c) R - c R' + \ln(e^{R} + e^{R'}). \label{eq:ce_loss}
\end{equation}
Note that when response $y$ is preferred over $y'$, the preference indicator $c$ is $0$ and vice versa.

\subsection{Epistemic Neural Network} \label{sec:ENN}

We use epistemic neural networks (ENNs) to model epistemic uncertainty about reward \citep{osband2023epistemic}.  Given the dataset $\data$, ENN parameters are obtained by minimizing the loss function
\begin{equation}\label{eq:enn_loss}
\mathcal{L}_{\rm ENN}(\theta | \data)  =  \lambda \|\theta - \tilde \theta\|_2  + \int_{z \in \mathcal{Z}} p_z(dz) \mathcal{L}(\theta| \data, z), 
\end{equation}
where $p_z$ is the epistemic index reference distribution, $\tilde \theta$ is the initial parameter vector, and 
\begin{equation*}
\mathcal{L}(\theta| \data, z) =  \sum_{(x, y, y', c) \in \data} \mathrm{ce}(r_\theta(x,y|z), r_\theta(x,y'|z),c).
\end{equation*}
To interpret these objects, note that with $z$ sampled from $p_z$, the reward function $r_{\tilde{\theta}}(\cdot|z)$ represents a sample from a prior distribution over reward functions.  In the loss function $\mathcal{L}_{\rm ENN}$, regularizing toward $\tilde{\theta}$ serves to maintain a suitable degree of diversity across epistemic indices after training.

\subsection{Training}

To train each reward model, we maintain a replay buffer and apply a stochastic gradient descent (SGD) algorithm with respect to loss functions described in Section \ref{sec:point} and \ref{sec:ENN}.  In particular, at the end of each epoch of interaction, over which the agent transmits $B$ queries and receives $B$ bits of feedback, the agent inserts the resulting $B$ data points into a FIFO replay buffer of capacity $C$.  Then, SGD steps are applied with random minibatches from the replay buffer, with step-sizes adapted by ADAM.The reward model that has been trained is employed to determine the queries formulated in the subsequent epoch.

\section{Exploration Algorithms} \label{sec:exploration-algorithms}

We now describe the set of exploration algorithms used in our empirical study.

\subsection{Passive Exploration}

Current RLHF systems typically explore passively, selecting response pairs according to Algorithm \ref{alg:passive}.  This algorithm takes a prompt $x$ and a language model $\pi$ as inputs.  The language model encodes a distribution $\pi(\cdot|x)$ from which it samples responses.  The algorithm returns two responses sampled by the language model.
\begin{algorithm}
\caption{passive exploration}
\label{alg:passive}
\textbf{input:} $x$, $\pi$
\begin{algorithmic}[1]
\STATE sample response $y \sim \pi(\cdot|x)$
\REPEAT
\STATE sample response $y' \sim \pi(\cdot|x)$
\UNTIL $y' \neq y$
\end{algorithmic}
\textbf{return} $y,y'$
\end{algorithm}

\subsection{Active Exploration with a Point Estimate}

When selecting a pair of responses, the agent can make use of a reward model that has been trained on feedback to all or some past queries.  Passive exploration forgoes this opportunity.  We now consider Boltzmann exploration, which makes use of a point estimate reward model, which assigns a reward $r(x,y)$ to each prompt-response pair.  This constitutes a form of active exploration: responses are tailored based on past feedback, with an aim to gather more useful future feedback than passive exploration.

As presented in Algorithm \ref{alg:Boltzmann}, in addition to the inputs $x$ and $\pi$ used for passive exploration, Boltzmann exploration requires a point estimate reward model $r$.  Further, there are two hyperparameters: a temperature $\tau$ and a response set cardinality $N$.  The language model generates $N$ responses, and two are sampled from a Boltzmann distribution with exponent $r(x,\tilde{y}_n) / \tau$ assigned to each $n$th response $\tilde{y}_n$.

\begin{algorithm}
\caption{Boltzmann}
\label{alg:Boltzmann}
\textbf{input:} $x$, $\pi$, $r$ \\
\textbf{hyperparams:} $\tau$, $N$
\begin{algorithmic}[1]
\STATE sample responses $\tilde{y}_1,\ldots,\tilde{y}_N \sim \pi(\cdot|x)$
\STATE probs $q_n = \frac{\exp(r(x,\tilde{y}_n) / \tau)}{\sum_{n'=1}^N \exp(r(x,\tilde{y}_{n'}) / \tau)}$, $\forall n$
\STATE sample without replacement $i,i' \sim q$
\end{algorithmic}
\textbf{return} $y_i,y_{i'}$
\end{algorithm}

Note that this algorithm recovers passive exploration as the temperature $\tau$ grows.  On the other hand, as $\tau$ vanishes, Boltzmann exploration tends to select responses that are optimal or nearly so.  One could also consider a generalization of the algorithm that uses two different temperatures $\tau_1$ and $\tau_2$ to select the two responses.  Then, for example, as $\tau_1$ vanishes and $\tau_2$ grows, the first response becomes optimal whereas the second is sampled uniformly.  In our experimental work, we have not found use of separate temperatures to improve performance.  Further, we have found Algorithm \ref{alg:Boltzmann} to offer the best performance among many alternatives that take the same inputs.  This suggests that Boltzmann exploration selects responses about as well as one can hope for based on a point estimate reward model.

\subsection{Active Exploration with an ENN}

We next consider algorithms that use an ENN reward model, for which the reward $r(x,y|z)$ assigned to each prompt-response pair depends additionally on an epistemic index.  As discussed in Section \ref{sec:ENN}, the ENN is characterized by the reward model $r$ and a reference distribution $p$.  For fixed $x$ and $y$, by sampling multiple epistemic indices from $p$, reward uncertainty can be ascertained from the variance among these samples.

Infomax (Algorithm \ref{alg:infomax}) takes an ENN reward model as input.  Like Boltzmann exploration (Algorithm \ref{alg:Boltzmann}), infomax begins with the language model generating $N$ responses.  Then, $M$ epistemic indices are sampled from $p$.  For each pair of responses and each epistemic index, the ENN assigns a probability to the event that a random human rater prefers the first response over the second.  Infomax assesses uncertainty about this probability by calculating a sample variance across the $M$ epistemic indices.  Then, the algorithm selects the pair of responses to maximize uncertainty.  Intuitively, this can be thought of as maximizing a measure of feedback informativeness.

\begin{algorithm}
\caption{infomax}
\label{alg:infomax}
\textbf{input:} $x$, $\pi$, $(r, p)$ \\
\textbf{hyperparams:} $N, M$
\begin{algorithmic}[1]
\STATE sample responses $\tilde{y}_1,\ldots,\tilde{y}_N \sim \pi(\cdot|x)$
\STATE sample indices $z_1,\ldots, z_M \sim p$
\STATE rewards $R_{n,m} = r(x,\tilde{y}_n|z_m)$, $\forall m, n$
\STATE pref probs $P_{n,n',m} = \frac{R_{n,m}}{(R_{n,m} + R_{n',m})}$, $\forall m, n,n'$
\STATE means $\mu_{n,n'} = \frac{\sum_m P_{n,n',m}}{M}$, $\forall n,n'$
\STATE vars $\sigma^2_{n,n'} = \frac{\sum_m (P_{n,n',m} - \mu_{n,n',m})^2}{M-1}$, $\forall n,n'$
\item $(i,i') \in \argmax_{n,n'} \sigma^2_{n,n'}$
\end{algorithmic}
\textbf{return} $y_i,y_{i'}$
\end{algorithm}

A possible limitation of infomax is that the algorithm invests in seeking information about rewards whether or not that information is useful to selecting the best responses.  For example, infomax can invest in refining an estimate of reward assigned to a response that has already been determined based on previous feedback to be a poor choice.  Double Thompson sampling \cite{NIPS2016_9de6d14f}, on the other hand, tends to focus more on queries that are helpful in identifying the best responses.  As we will see in Section \ref{se:results}, double TS improves on the performance of infomax, as well as Boltzmann exploration.

Intuitively, double TS (Algorithm \ref{alg:DTS}) aims to select two responses that each have some chance of being optimal.  Like Algorithms \ref{alg:Boltzmann} and \ref{alg:infomax}, we begin by sampling $N$ responses.  Then, two among these $N$ responses are selected by sampling two epistemic indices from $p$ and maximizing across rewards prescribed by each.  In the event that samples are identical, the second response is resampled until it differs.  If there is no difference after $K$ iterations, the second response is instead sampled uniformly.

\begin{algorithm}
\caption{double Thompson sampling}
\label{alg:DTS}
\textbf{input:} $x$, $\pi$, $(r,p)$ \\
\textbf{hyperparams:} $N$, $K$
\begin{algorithmic}[1]
\STATE sample responses $\tilde{y}_1,\ldots,\tilde{y}_N \sim \pi(\cdot|x)$
\STATE sample index $z \sim p$
\STATE select response $i \in \argmax_{n} r(x,\tilde{y}_n|z)$
\REPEAT
\STATE sample index $z' \sim p$
\STATE select response $i' \in \argmax_{n} r(x,\tilde{y}_n|z')$
\STATE after $K$ tries, instead sample $i' \sim \mathrm{unif}(1,\ldots,N)$
\UNTIL $i' \neq i$
\end{algorithmic}
\textbf{return} $y_i,y_{i'}$
\end{algorithm}

\section{Empirical Results}
\label{se:results}

In our experiments, at the start of each epoch of interaction, each agents receives a batch of $B=32$ prompts and then, for each prompt, generates a pair of responses to form a query.  Each agent's $B=32$ queries are submitted to the preference simulator, yielding $B=32$ bits of feedback.  Each agent inserts its batch of $B=32$ data points into its replay buffer.  The replay buffers are first-in-first-out (FIFO) buffer, with a maximum capacity of $C=3200$ data points. In other words, replay buffer holds preference data from a maximum of $100$ most recent epochs. At the end of each epoch, each agent updates its reward model as discussed in Section \ref{sec:reward-learning-algorithms}.

Recall that each exploration algorithm selects each pair of responses from $N$ candidates sampled by \smallllm{}.  In our experiments, we set $N=100$.  Performance is assessed in terms of win rate relative to \smallllm{} on $2048$ out-of-sample Anthropic Helpfulness base eval prompts, as explained in Section \ref{sec:pipeline}.  Each response selected in this assessment is chosen to score highest among $N=100$ candidates sampled by \smallllm{} according to the agent's reward model. Note that we use $N=100$ responses both in our training and assement piplelines.

For a singular point estimate, we employ a feedforward multilayer perceptron (MLP) comprising two hidden layers, with $128$ hidden units in each layer. As an ENN architecture, we utilize a collection of $S=10$ MLPs, referring to each individual MLP as a particle. Each particle of ensemble consists of two $128$ unit hidden layers. The reference distribution $p_z$ is defined as the uniform distribution on $\{1, 2, \ldots, S\}$. When selecting an epistemic index $z$ sampled from $\mathrm{Unif}({ 1, 2, \ldots, S})$, particle $z$ is utilized to produce the output for that specific index $z$.  The ENN loss function presented in Section \ref{sec:ENN} maintains diversity across particles by regularizing each toward initial parameters. 

For the \texttt{Boltzmann} exploration scheme, we swept over several temperatures and found that small temperatures produced best results. A similar level of performance was achieved by a variant of Boltzmann scheme that selects one of the response greedily and the second response using Boltzmann. More details can be found in Appendix \ref{app:point}.

In the case of \texttt{infomax}, we used $30$ epistemic indices to compute means and variances.  For \texttt{double TS} agent, we set the maximum number of attempts at producing a distinct second response to $K=30$.

Appendix \ref{app:hypers} presents further detail on our hyperparameter selection process.

% By default, we use a learning rate of $1e-5$ for all the agents and do $10$ sgd steps after each epoch by sampling $10$ minbatches from the replay buffer.  

\subsection{Assessment of Exploration Algorithms}

\begin{figure}[!ht]
    \centering
    \includegraphics[width=0.6\columnwidth]{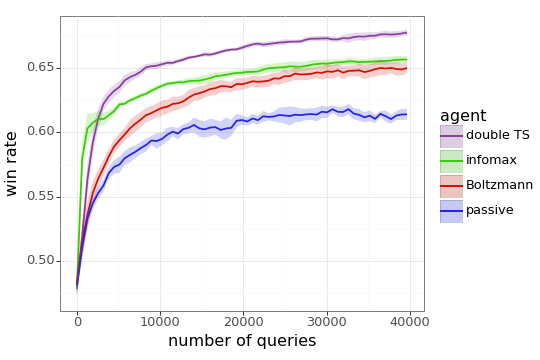}
    \caption{Performance with passive, \texttt{Boltzmann}, \texttt{infomax} and \texttt{double TS} exploration algorithms. We can see that active exploration leads to much better levels of performance with the same amount of data. \texttt{double TS} exploration scheme leads to the best level of performance.}
    \label{fig:passive_vs_best}
\end{figure}

Figure \ref{fig:passive_vs_best} plots win rates of each agent across different numbers of epochs of interactions.  The results, obtained by averaging across $5$ random seeds, clearly demonstrate that active exploration accelerates learning and results in higher win rates. Notably, the \texttt{double TS} agent emerges as the top performer.

We observe that \texttt{infomax} performs very well over early epochs but later falls far short of \texttt{double TS}.  This divergence may be due to \texttt{infomax}'s inclination to seek information, irrespective of whether that information is helpful in desirable responses.

Each of the performance curves in Figure \ref{fig:passive_vs_best} appears to converge, while one would hope for continued improvement as the volume of human interaction grows.  Reward model capacity -- which can be thought of loosely as the effective number of parameters learned from feedback -- gates the degree of improvement.  For any capacity, one would expect convergence as the number of queries grows.  Increasing the capacity enables further improvement at the cost of increased computation.  This relates to the notion explained by \citet{pmlr-v139-arumugam21a} that it is beneficial to moderate the complexity of a learning target based on the duration over which an agent expects to explore.

\subsection{Scaling with the Volume of Feedback}

\begin{figure}[!ht]
\centering
\includegraphics[width=0.6\columnwidth]{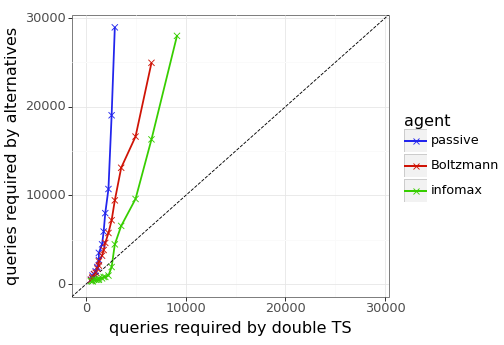}
\renewcommand\thefigure{\ref{fig:active-vs-passive}}
\caption{Queries required by double TS versus alternatives to attain various levels of performance.}
\end{figure}
\addtocounter{figure}{-1}

Figure \ref{fig:active-vs-passive}, reproduced from Section \ref{se:introduction} for convenience, plots the number of queries required by alternatives to match the performance of {\tt double TS}, which we found to be most efficient among exploration algorithms we considered.  While the plots are not conclusive, we discern that they are concave.  Suppose we measure the advantage of efficient exploration in terms of the percentage reduction in data required to attain any given level of performance.  Concavity of the plots in Figure \ref{fig:active-vs-passive} implies that, as the scale of human feedback data grows, so does the advantage afforded by efficient exploration.  For the level of performance attained by $30,000$ passive queries, {\tt double TS} reduces data requirements by an order of magnitude.  An alluring possibility is that, as the number of interactions grow to billions, efficient exploration may offer a multiplier effect reaching several orders of magnitude.  This has the potential to accelerate by decades the attainment of superhuman creativity.

\subsection{Quality of Uncertainty Estimates} \label{sec:joint_nll}

\texttt{Boltzmann} exploration performed best among algorithms we tried that select queries based on a point estimate reward model.  The large improvement demonstrated by \texttt{double TS} is enabled by uncertainty estimates offered by our ENN reward model.

\begin{figure}
    \centering
    \begin{minipage}{.45\columnwidth}
    \includegraphics[width=0.6\columnwidth]{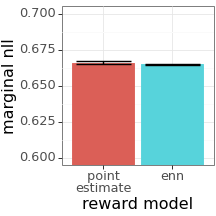}
    \captionof{figure}{Marginal nll}
    \label{fig:marginal-nll}
    \end{minipage}%
    \hspace{0.5cm}%
    \begin{minipage}{.45\columnwidth}
    \includegraphics[width=0.6\columnwidth]{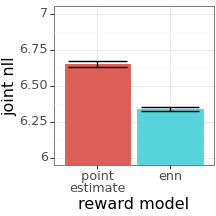}
    \captionof{figure}{Dyadic joint nll}
    \label{fig:joint-nll}
    \end{minipage}
\end{figure}

The quality of uncertainty estimates can be assessed in terms of dyadic joint negative-log loss (NLL) \cite{osband2022evaluating}, using preference probabilities. Figures \ref{fig:marginal-nll} and \ref{fig:joint-nll} plot marginal and dyadic joint NLL for our point estimate and ENN reward models, each trained on $40,000$ queries.  These plots indicate that, while both reward models render similar marginal NLL, the ENN reward model offers highly favorable dyadic joint NLL.  This serves as a sanity check that our ENN reward model indeed produces meaningful uncertainty estimates.

We also used dyadic joint NLL to guide hyperparameter selection for our point estimate and ENN reward models used by our exploration algorithms.  In particular, we swept over candidates for learning rate, training the agent over multiple epochs to identify learning rate the minimize dyadic joint NLL.

\subsection{The Life of a Prompt}

Our results indicate that double TS tends to converge on better responses than the alternatives.  To understand more concretely how this occurs, let us study the evolution of rewards that models assign to responses to a specific prompt.  To simplify this investigation, we will only compare double TS against Boltzmann exploration.

Recall that we found Boltzmann exploration to be the top performer among algorithms that base decisions on a point estimate reward model.  Double TS, on the other hand, makes use of uncertainty estimates offered by an ENN reward model.  We will examine estimates associated with a single prompt and two responses, selected from the eval data set.  The first is the response that double TS arrives at, while the second is the response that Boltzmann exploration arrives at.  The human feedback simulator indicates preference for the first prompt $57.5\%$ of the time.

\begin{figure}[!ht]
    \centering
    \includegraphics[width=0.6\columnwidth]{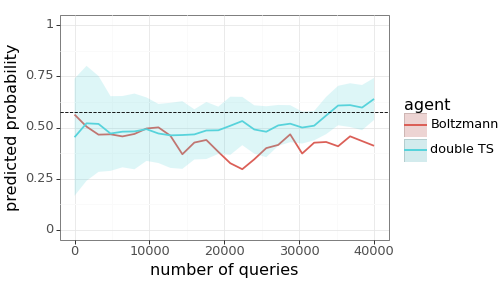}
    \caption{For a particular prompt, the dotted line indicates the probability that the simulator expresses preference for one response over another.  Uncertainty estimates enable double TS to recover from an inaccurate prediction where Boltzmann exploration does not.}
    \label{fig:life_of_prompt}
\end{figure}

Figure \ref{fig:life_of_prompt} plots the prediction supplied by each reward model of the probability that the first response is preferred.  The horizontal dotted line expresses the probability of $0.575$ with which the feedback simulator expresses preference for the first response.  The predictions evolve as the reward models learn from queries.  After 40,000 queries, double TS arrives at a prediction that is greater than one-half, expressing preference for the first response.  Boltzmann exploration, on the other hand, expresses preference for the second with a prediction that is less than one-half.

Also displayed in the figure is the two-standard-deviation confidence interval based on uncertainty expressed by the ENN reward model.  Though double TS at some points predicts less than one-half, the upper limit of its confidence interval remains greater than one-half.  Hence, it remains uncertain about which is the better response.  In resolving this uncertainty, it recovers and arrives at a prediction greater than one-half.  Boltzmann exploration, on the other hand, is not guided by uncertainty estimates and thus does not recover from its erroneous prediction.

\section{Closing Remarks}

To our knowledge, the results we have presented represent the first to demonstrate substantial benefits of active exploration in tuning large language models.  That being said, there is much room for further work in this area.  To conclude this paper, we discuss several important research directions.

Our experiments made use of a particularly simple ENN architecture comprised of an ensemble of MLPs.  As demonstrated in \cite{osband2023epistemic}, alternative architectures strike a more effective tradeoff between computational requirements and quality of uncertainty estimates.  Further, instead of designing ENNs based on the MLP, it may be possible to improve performance, especially as the amount of human feedback data grows, by basing ENN designs on transformer architectures.

Another limitation of our reward model architectures is that each is only a ``head'' that takes the last-layer embedding of an LLM as input.  Performance can be improved by also tuning the LLM torso.  While advantages afforded by efficient exploration should extend, identifying the most effective architectures and algorithms for exploring while tuning more of the LLM remains for future work.

Finally, efficient exploration of multiturn dialog presents an interesting and important direction for future research.  In this paper, we viewed exploration as a means to quickly identifying a response deemed desirable in isolation.  In multiturn dialog, responses may be chosen instead because of how they shape subsequent interactions.  The subject of deep exploration addresses how an agent can efficiently identify effective responses that make up sequential interactions \cite{osband2016deep,osband2019deep}.  Leveraging deep exploration algorithms to improve dialog remains a challenge.

%% file: arxiv.bbl
\begin{thebibliography}{40}
\providecommand{\natexlab}[1]{#1}
\providecommand{\url}[1]{\texttt{#1}}
\expandafter\ifx\csname urlstyle\endcsname\relax
  \providecommand{\doi}[1]{doi: #1}\else
  \providecommand{\doi}{doi: \begingroup \urlstyle{rm}\Url}\fi

\bibitem[Anil et~al.(2023)Anil, Dai, Firat, Johnson, Lepikhin, Passos, Shakeri,
  Taropa, Bailey, Chen, Chu, Clark, Shafey, Huang, Meier-Hellstern, Mishra,
  Moreira, Omernick, Robinson, Ruder, Tay, Xiao, Xu, Zhang, Abrego, Ahn,
  Austin, Barham, Botha, Bradbury, Brahma, Brooks, Catasta, Cheng, Cherry,
  Choquette-Choo, Chowdhery, Crepy, Dave, Dehghani, Dev, Devlin, Díaz, Du,
  Dyer, Feinberg, Feng, Fienber, Freitag, Garcia, Gehrmann, Gonzalez, Gur-Ari,
  Hand, Hashemi, Hou, Howland, Hu, Hui, Hurwitz, Isard, Ittycheriah, Jagielski,
  Jia, Kenealy, Krikun, Kudugunta, Lan, Lee, Lee, Li, Li, Li, Li, Li, Lim, Lin,
  Liu, Liu, Maggioni, Mahendru, Maynez, Misra, Moussalem, Nado, Nham, Ni,
  Nystrom, Parrish, Pellat, Polacek, Polozov, Pope, Qiao, Reif, Richter, Riley,
  Ros, Roy, Saeta, Samuel, Shelby, Slone, Smilkov, So, Sohn, Tokumine, Valter,
  Vasudevan, Vodrahalli, Wang, Wang, Wang, Wang, Wieting, Wu, Xu, Xu, Xue, Yin,
  Yu, Zhang, Zheng, Zheng, Zhou, Zhou, Petrov, and Wu]{anil2023palm}
R.~Anil, A.~M. Dai, O.~Firat, M.~Johnson, D.~Lepikhin, A.~Passos, S.~Shakeri,
  E.~Taropa, P.~Bailey, Z.~Chen, E.~Chu, J.~H. Clark, L.~E. Shafey, Y.~Huang,
  K.~Meier-Hellstern, G.~Mishra, E.~Moreira, M.~Omernick, K.~Robinson,
  S.~Ruder, Y.~Tay, K.~Xiao, Y.~Xu, Y.~Zhang, G.~H. Abrego, J.~Ahn, J.~Austin,
  P.~Barham, J.~Botha, J.~Bradbury, S.~Brahma, K.~Brooks, M.~Catasta, Y.~Cheng,
  C.~Cherry, C.~A. Choquette-Choo, A.~Chowdhery, C.~Crepy, S.~Dave,
  M.~Dehghani, S.~Dev, J.~Devlin, M.~Díaz, N.~Du, E.~Dyer, V.~Feinberg,
  F.~Feng, V.~Fienber, M.~Freitag, X.~Garcia, S.~Gehrmann, L.~Gonzalez,
  G.~Gur-Ari, S.~Hand, H.~Hashemi, L.~Hou, J.~Howland, A.~Hu, J.~Hui,
  J.~Hurwitz, M.~Isard, A.~Ittycheriah, M.~Jagielski, W.~Jia, K.~Kenealy,
  M.~Krikun, S.~Kudugunta, C.~Lan, K.~Lee, B.~Lee, E.~Li, M.~Li, W.~Li, Y.~Li,
  J.~Li, H.~Lim, H.~Lin, Z.~Liu, F.~Liu, M.~Maggioni, A.~Mahendru, J.~Maynez,
  V.~Misra, M.~Moussalem, Z.~Nado, J.~Nham, E.~Ni, A.~Nystrom, A.~Parrish,
  M.~Pellat, M.~Polacek, A.~Polozov, R.~Pope, S.~Qiao, E.~Reif, B.~Richter,
  P.~Riley, A.~C. Ros, A.~Roy, B.~Saeta, R.~Samuel, R.~Shelby, A.~Slone,
  D.~Smilkov, D.~R. So, D.~Sohn, S.~Tokumine, D.~Valter, V.~Vasudevan,
  K.~Vodrahalli, X.~Wang, P.~Wang, Z.~Wang, T.~Wang, J.~Wieting, Y.~Wu, K.~Xu,
  Y.~Xu, L.~Xue, P.~Yin, J.~Yu, Q.~Zhang, S.~Zheng, C.~Zheng, W.~Zhou, D.~Zhou,
  S.~Petrov, and Y.~Wu.
\newblock Palm 2 technical report, 2023.

\bibitem[Arumugam and Van~Roy(2021)]{pmlr-v139-arumugam21a}
D.~Arumugam and B.~Van~Roy.
\newblock Deciding what to learn: A rate-distortion approach.
\newblock In \emph{Proceedings of the 38th International Conference on Machine
  Learning}, pages 373--382, 2021.

\bibitem[Badia et~al.(2020)Badia, Sprechmann, Vitvitskyi, Guo, Piot,
  Kapturowski, Tieleman, Arjovsky, Pritzel, Bolt, et~al.]{badia2020never}
A.~P. Badia, P.~Sprechmann, A.~Vitvitskyi, D.~Guo, B.~Piot, S.~Kapturowski,
  O.~Tieleman, M.~Arjovsky, A.~Pritzel, A.~Bolt, et~al.
\newblock Never give up: Learning directed exploration strategies.
\newblock \emph{arXiv preprint arXiv:2002.06038}, 2020.

\bibitem[Bai et~al.(2022)Bai, Jones, Ndousse, Askell, Chen, DasSarma, Drain,
  Fort, Ganguli, Henighan, et~al.]{bai2022training}
Y.~Bai, A.~Jones, K.~Ndousse, A.~Askell, A.~Chen, N.~DasSarma, D.~Drain,
  S.~Fort, D.~Ganguli, T.~Henighan, et~al.
\newblock Training a helpful and harmless assistant with reinforcement learning
  from human feedback.
\newblock \emph{arXiv preprint arXiv:2204.05862}, 2022.

\bibitem[Bellemare et~al.(2016)Bellemare, Srinivasan, Ostrovski, Schaul,
  Saxton, and Munos]{bellemare2016unifying}
M.~Bellemare, S.~Srinivasan, G.~Ostrovski, T.~Schaul, D.~Saxton, and R.~Munos.
\newblock Unifying count-based exploration and intrinsic motivation.
\newblock \emph{Advances in neural information processing systems}, 29, 2016.

\bibitem[Bradley and Terry(1952)]{bradley1952rank}
R.~A. Bradley and M.~E. Terry.
\newblock Rank {A}nalysis of {I}ncomplete {B}lock {D}esigns: {I}. {T}he
  {M}ethod of {P}aired {C}omparisons.
\newblock \emph{Biometrika}, 39\penalty0 (3/4):\penalty0 324--345, 1952.

\bibitem[Burda et~al.(2018)Burda, Edwards, Storkey, and
  Klimov]{burda2018exploration}
Y.~Burda, H.~Edwards, A.~Storkey, and O.~Klimov.
\newblock Exploration by random network distillation.
\newblock \emph{arXiv preprint arXiv:1810.12894}, 2018.

\bibitem[Dud{\'\i}k et~al.(2015)Dud{\'\i}k, Hofmann, Schapire, Slivkins, and
  Zoghi]{dudik2015contextual}
M.~Dud{\'\i}k, K.~Hofmann, R.~E. Schapire, A.~Slivkins, and M.~Zoghi.
\newblock Contextual dueling bandits.
\newblock In \emph{Conference on Learning Theory}, pages 563--587. PMLR, 2015.

\bibitem[Dwaracherla et~al.(2020)Dwaracherla, Lu, Ibrahimi, Osband, Wen, and
  Van~Roy]{dwaracherla2020hypermodels}
V.~Dwaracherla, X.~Lu, M.~Ibrahimi, I.~Osband, Z.~Wen, and B.~Van~Roy.
\newblock Hypermodels for exploration.
\newblock In \emph{International Conference on Learning Representations}, 2020.

\bibitem[Glaese et~al.(2022)Glaese, McAleese, Trebacz, Aslanides, Firoiu,
  Ewalds, Rauh, Weidinger, Chadwick, Thacker, et~al.]{glaese2022improving}
A.~Glaese, N.~McAleese, M.~Trebacz, J.~Aslanides, V.~Firoiu, T.~Ewalds,
  M.~Rauh, L.~Weidinger, M.~Chadwick, P.~Thacker, et~al.
\newblock Improving alignment of dialogue agents via targeted human judgements.
\newblock \emph{arXiv preprint arXiv:2209.14375}, 2022.

\bibitem[Glorot and Bengio(2010)]{glorot2010understanding}
X.~Glorot and Y.~Bengio.
\newblock Understanding the difficulty of training deep feedforward neural
  networks.
\newblock In \emph{Proceedings of the thirteenth international conference on
  artificial intelligence and statistics}, pages 249--256. JMLR Workshop and
  Conference Proceedings, 2010.

\bibitem[Hennig and Schuler(2012)]{hennig2012entropy}
P.~Hennig and C.~J. Schuler.
\newblock Entropy search for information-efficient global optimization.
\newblock \emph{Journal of Machine Learning Research}, 13\penalty0 (6), 2012.

\bibitem[Hoffmann et~al.(2022)Hoffmann, Borgeaud, Mensch, Buchatskaya, Cai,
  Rutherford, de~Las~Casas, Hendricks, Welbl, Clark, Hennigan, Noland,
  Millican, van~den Driessche, Damoc, Guy, Osindero, Simonyan, Elsen, Vinyals,
  Rae, and Sifre]{NEURIPS2022_c1e2faff}
J.~Hoffmann, S.~Borgeaud, A.~Mensch, E.~Buchatskaya, T.~Cai, E.~Rutherford,
  D.~de~Las~Casas, L.~A. Hendricks, J.~Welbl, A.~Clark, T.~Hennigan, E.~Noland,
  K.~Millican, G.~van~den Driessche, B.~Damoc, A.~Guy, S.~Osindero,
  K.~Simonyan, E.~Elsen, O.~Vinyals, J.~Rae, and L.~Sifre.
\newblock An empirical analysis of compute-optimal large language model
  training.
\newblock In S.~Koyejo, S.~Mohamed, A.~Agarwal, D.~Belgrave, K.~Cho, and A.~Oh,
  editors, \emph{Advances in Neural Information Processing Systems}, volume~35,
  pages 30016--30030. Curran Associates, Inc., 2022.

\bibitem[Houthooft et~al.(2016)Houthooft, Chen, Chen, Duan, Schulman, De~Turck,
  and Abbeel]{Houthooft2016}
R.~Houthooft, X.~Chen, X.~Chen, Y.~Duan, J.~Schulman, F.~De~Turck, and
  P.~Abbeel.
\newblock Vime: Variational information maximizing exploration.
\newblock In D.~Lee, M.~Sugiyama, U.~Luxburg, I.~Guyon, and R.~Garnett,
  editors, \emph{Advances in Neural Information Processing Systems}, volume~29.
  Curran Associates, Inc., 2016.

\bibitem[Lattimore and Szepesv{\'a}ri(2020)]{lattimore2020bandit}
T.~Lattimore and C.~Szepesv{\'a}ri.
\newblock \emph{{B}andit {A}lgorithms}.
\newblock Cambridge University Press, 2020.

\bibitem[Lu and Van~Roy(2017)]{lu2017ensemble}
X.~Lu and B.~Van~Roy.
\newblock {E}nsemble {S}ampling.
\newblock In \emph{Proceedings of the 31st International Conference on Neural
  Information Processing Systems}, pages 3260--3268, 2017.

\bibitem[MacKay(1992)]{mackay1992information}
D.~J. MacKay.
\newblock Information-based objective functions for active data selection.
\newblock \emph{Neural computation}, 4\penalty0 (4):\penalty0 590--604, 1992.

\bibitem[OpenAI(2022)]{openai2022chatgpt}
OpenAI.
\newblock {ChatGPT}: {O}ptimizing {L}anguage {M}odels for {D}ialogue, 2022.
\newblock URL \url{https://openai.com/blog/chatgpt/}.

\bibitem[OpenAI(2023)]{openai2023gpt4}
OpenAI.
\newblock {GPT-4} {T}echnical {R}eport.
\newblock Technical report, OpenAI, 2023.

\bibitem[Osband et~al.(2016)Osband, Blundell, Pritzel, and
  Van~Roy]{osband2016deep}
I.~Osband, C.~Blundell, A.~Pritzel, and B.~Van~Roy.
\newblock Deep exploration via bootstrapped {DQN}.
\newblock In D.~Lee, M.~Sugiyama, U.~Luxburg, I.~Guyon, and R.~Garnett,
  editors, \emph{Advances in Neural Information Processing Systems}, volume~29.
  Curran Associates, Inc., 2016.

\bibitem[Osband et~al.(2019)Osband, Van~Roy, Russo, and Wen]{osband2019deep}
I.~Osband, B.~Van~Roy, D.~J. Russo, and Z.~Wen.
\newblock Deep exploration via randomized value functions.
\newblock \emph{Journal of Machine Learning Research}, 20\penalty0
  (124):\penalty0 1--62, 2019.

\bibitem[Osband et~al.(2022)Osband, Wen, Asghari, Dwaracherla, Lu, and
  Van~Roy]{osband2022evaluating}
I.~Osband, Z.~Wen, S.~M. Asghari, V.~Dwaracherla, X.~Lu, and B.~Van~Roy.
\newblock Evaluating high-order predictive distributions in deep learning.
\newblock In \emph{The 38th Conference on Uncertainty in Artificial
  Intelligence}, 2022.

\bibitem[Osband et~al.(2023{\natexlab{a}})Osband, Wen, Asghari, Dwaracherla,
  Ibrahimi, Lu, and Van~Roy]{osband2023epistemic}
I.~Osband, Z.~Wen, M.~Asghari, V.~Dwaracherla, M.~Ibrahimi, X.~Lu, and
  B.~Van~Roy.
\newblock Epistemic neural networks.
\newblock \emph{Advances in Neural Information Processing Systems}, 34,
  2023{\natexlab{a}}.

\bibitem[Osband et~al.(2023{\natexlab{b}})Osband, Wen, Asghari, Dwaracherla,
  Ibrahimi, Lu, and Van~Roy]{pmlr-v216-osband23a}
I.~Osband, Z.~Wen, S.~M. Asghari, V.~Dwaracherla, M.~Ibrahimi, X.~Lu, and
  B.~Van~Roy.
\newblock Approximate {T}hompson sampling via epistemic neural networks.
\newblock In R.~J. Evans and I.~Shpitser, editors, \emph{Proceedings of the
  Thirty-Ninth Conference on Uncertainty in Artificial Intelligence}, volume
  216 of \emph{Proceedings of Machine Learning Research}, pages 1586--1595.
  PMLR, 31 Jul--04 Aug 2023{\natexlab{b}}.

\bibitem[Ostrovski et~al.(2017)Ostrovski, Bellemare, Oord, and
  Munos]{ostrovski2017count}
G.~Ostrovski, M.~G. Bellemare, A.~Oord, and R.~Munos.
\newblock Count-based exploration with neural density models.
\newblock In \emph{International conference on machine learning}, pages
  2721--2730. PMLR, 2017.

\bibitem[Ouyang et~al.(2022)Ouyang, Wu, Jiang, Almeida, Wainwright, Mishkin,
  Zhang, Agarwal, Slama, Ray, Schulman, Hilton, Kelton, Miller, Simens, Askell,
  Welinder, Christiano, Leike, and Lowe]{ouyang2022training}
L.~Ouyang, J.~Wu, X.~Jiang, D.~Almeida, C.~Wainwright, P.~Mishkin, C.~Zhang,
  S.~Agarwal, K.~Slama, A.~Ray, J.~Schulman, J.~Hilton, F.~Kelton, L.~Miller,
  M.~Simens, A.~Askell, P.~Welinder, P.~F. Christiano, J.~Leike, and R.~Lowe.
\newblock Training language models to follow instructions with human feedback.
\newblock In S.~Koyejo, S.~Mohamed, A.~Agarwal, D.~Belgrave, K.~Cho, and A.~Oh,
  editors, \emph{Advances in Neural Information Processing Systems}, volume~35,
  pages 27730--27744. Curran Associates, Inc., 2022.

\bibitem[Riquelme et~al.(2018)Riquelme, Tucker, and Snoek]{riquelme2018deep}
C.~Riquelme, G.~Tucker, and J.~Snoek.
\newblock Deep {Bayesian} bandits showdown: An empirical comparison of
  {Bayesian} deep networks for {Thompson} sampling.
\newblock \emph{arXiv preprint arXiv:1802.09127}, 2018.

\bibitem[Russo and Van~Roy(2014)]{russo2014learning}
D.~Russo and B.~Van~Roy.
\newblock Learning to optimize via information-directed sampling.
\newblock \emph{Advances in Neural Information Processing Systems},
  27:\penalty0 1583--1591, 2014.

\bibitem[Russo et~al.(2018)Russo, Van~Roy, Kazerouni, Osband, and
  Wen]{russo2018tutorial}
D.~J. Russo, B.~Van~Roy, A.~Kazerouni, I.~Osband, and Z.~Wen.
\newblock {A} {T}utorial on {T}hompson {S}ampling.
\newblock \emph{Foundations and Trends{\textregistered} in Machine Learning},
  11\penalty0 (1):\penalty0 1--96, 2018.

\bibitem[Ryzhov et~al.(2012)Ryzhov, Powell, and Frazier]{ryzhov2012knowledge}
I.~O. Ryzhov, W.~B. Powell, and P.~I. Frazier.
\newblock The knowledge gradient algorithm for a general class of online
  learning problems.
\newblock \emph{Operations Research}, 60\penalty0 (1):\penalty0 180--195, 2012.

\bibitem[Sadigh et~al.(2018)Sadigh, Landolfi, Sastry, Seshia, and
  Dragan]{sadigh2018planning}
D.~Sadigh, N.~Landolfi, S.~S. Sastry, S.~A. Seshia, and A.~D. Dragan.
\newblock Planning for cars that coordinate with people: Leveraging effects on
  human actions for planning and active information gathering over human
  internal state.
\newblock \emph{Autonomous Robots (AURO)}, 42\penalty0 (7):\penalty0
  1405--1426, Oct. 2018.
\newblock ISSN 1573-7527.
\newblock \doi{10.1007/s10514-018-9746-1}.

\bibitem[Saha(2021)]{saha2021optimal}
A.~Saha.
\newblock Optimal algorithms for stochastic contextual preference bandits.
\newblock \emph{Advances in Neural Information Processing Systems},
  34:\penalty0 30050--30062, 2021.

\bibitem[Stiennon et~al.(2020)Stiennon, Ouyang, Wu, Ziegler, Lowe, Voss,
  Radford, Amodei, and Christiano]{stiennon2020learning}
N.~Stiennon, L.~Ouyang, J.~Wu, D.~Ziegler, R.~Lowe, C.~Voss, A.~Radford,
  D.~Amodei, and P.~F. Christiano.
\newblock Learning to summarize with human feedback.
\newblock \emph{Advances in Neural Information Processing Systems},
  33:\penalty0 3008--3021, 2020.

\bibitem[Sun et~al.(2011)Sun, Gomez, and Schmidhuber]{Sun2011}
Y.~Sun, F.~Gomez, and J.~Schmidhuber.
\newblock Planning to be surprised: Optimal {B}ayesian exploration in dynamic
  environments.
\newblock In J.~Schmidhuber, K.~R. Th{\'o}risson, and M.~Looks, editors,
  \emph{Artificial General Intelligence}, pages 41--51, Berlin, Heidelberg,
  2011. Springer Berlin Heidelberg.

\bibitem[Team et~al.(2023)Team, Anil, Borgeaud, Wu, Alayrac, Yu, Soricut,
  Schalkwyk, Dai, Hauth, et~al.]{geminiteam2023gemini}
G.~Team, R.~Anil, S.~Borgeaud, Y.~Wu, J.-B. Alayrac, J.~Yu, R.~Soricut,
  J.~Schalkwyk, A.~M. Dai, A.~Hauth, et~al.
\newblock Gemini: a family of highly capable multimodal models, 2023.

\bibitem[Thompson(1933)]{thompson1933likelihood}
W.~R. Thompson.
\newblock On the likelihood that one unknown probability exceeds another in
  view of the evidence of two samples.
\newblock \emph{Biometrika}, 25\penalty0 (3/4):\penalty0 285--294, 1933.

\bibitem[Wu and Liu(2016)]{NIPS2016_9de6d14f}
H.~Wu and X.~Liu.
\newblock Double {T}hompson sampling for dueling bandits.
\newblock In D.~Lee, M.~Sugiyama, U.~Luxburg, I.~Guyon, and R.~Garnett,
  editors, \emph{Advances in Neural Information Processing Systems}, volume~29.
  Curran Associates, Inc., 2016.

\bibitem[Yue et~al.(2012)Yue, Broder, Kleinberg, and Joachims]{yue2012k}
Y.~Yue, J.~Broder, R.~Kleinberg, and T.~Joachims.
\newblock The {$K$}-armed dueling bandits problem.
\newblock \emph{Journal of Computer and System Sciences}, 78\penalty0
  (5):\penalty0 1538--1556, 2012.

\bibitem[Zhang et~al.(2020)Zhang, Zhou, Li, and Gu]{zhang2020neural}
W.~Zhang, D.~Zhou, L.~Li, and Q.~Gu.
\newblock Neural {T}hompson sampling.
\newblock \emph{arXiv preprint arXiv:2010.00827}, 2020.

\bibitem[Zhou et~al.(2020)Zhou, Li, and Gu]{zhou2020neural}
D.~Zhou, L.~Li, and Q.~Gu.
\newblock Neural contextual bandits with ucb-based exploration.
\newblock In \emph{International Conference on Machine Learning}, pages
  11492--11502. PMLR, 2020.

\end{thebibliography}
